\pdfoutput=1

\documentclass[11pt]{article}

\usepackage{acl}

\usepackage{times}
\usepackage{latexsym}

\usepackage[T1]{fontenc}

\usepackage[utf8]{inputenc}

\usepackage{microtype}

\usepackage{colortbl}
\usepackage{times}
\usepackage{latexsym}
\usepackage{amsmath}
\usepackage{amssymb}
\usepackage{times}
\usepackage{latexsym}
\usepackage{xcolor}
\usepackage{bm}
\usepackage{multirow}
\usepackage{booktabs}
\usepackage{graphicx}
\usepackage{pifont}
\usepackage{tikz}
\usetikzlibrary{bayesnet,arrows,positioning,calc}

\DeclareMathOperator*{\maxB}{max}

\newcommand{\cmark}{\ding{51}}%
\newcommand{\xmark}{\ding{55}}%
\usepackage{fontawesome}
\usepackage{arydshln}

\title{Abductive Commonsense Reasoning\\ Exploiting Mutually Exclusive Explanations}

\author{Wenting Zhao \and Justin T. Chiu \and Claire Cardie  \and Alexander M. Rush\\
        Department of Computer Science \\ Cornell University \\ \texttt{\{wz346,jtc257,ctc9,arush\}@cornell.edu}}

\begin{document}
\maketitle
\begin{abstract}
    Abductive reasoning aims to find \textit{plausible} explanations for an event. This style of reasoning is critical for commonsense tasks where there are often multiple plausible explanations. Existing approaches for abductive reasoning in natural language processing (NLP) often rely on manually generated annotations for supervision; however, such annotations can be subjective and biased. Instead of using direct supervision, this work proposes an approach for abductive commonsense reasoning that exploits the fact that only a subset of explanations is correct for a given context. The method uses posterior regularization to enforce a mutual exclusion constraint, encouraging the model to learn the distinction between fluent explanations and plausible ones.
    We evaluate our approach on a diverse set of abductive reasoning datasets; experimental results show that our approach outperforms or is comparable to directly applying pretrained language models in a zero-shot manner and other knowledge-augmented zero-shot methods.
    
\end{abstract}
\section{Introduction}
\begin{figure}[t]
\centering
\includegraphics[width=\linewidth]{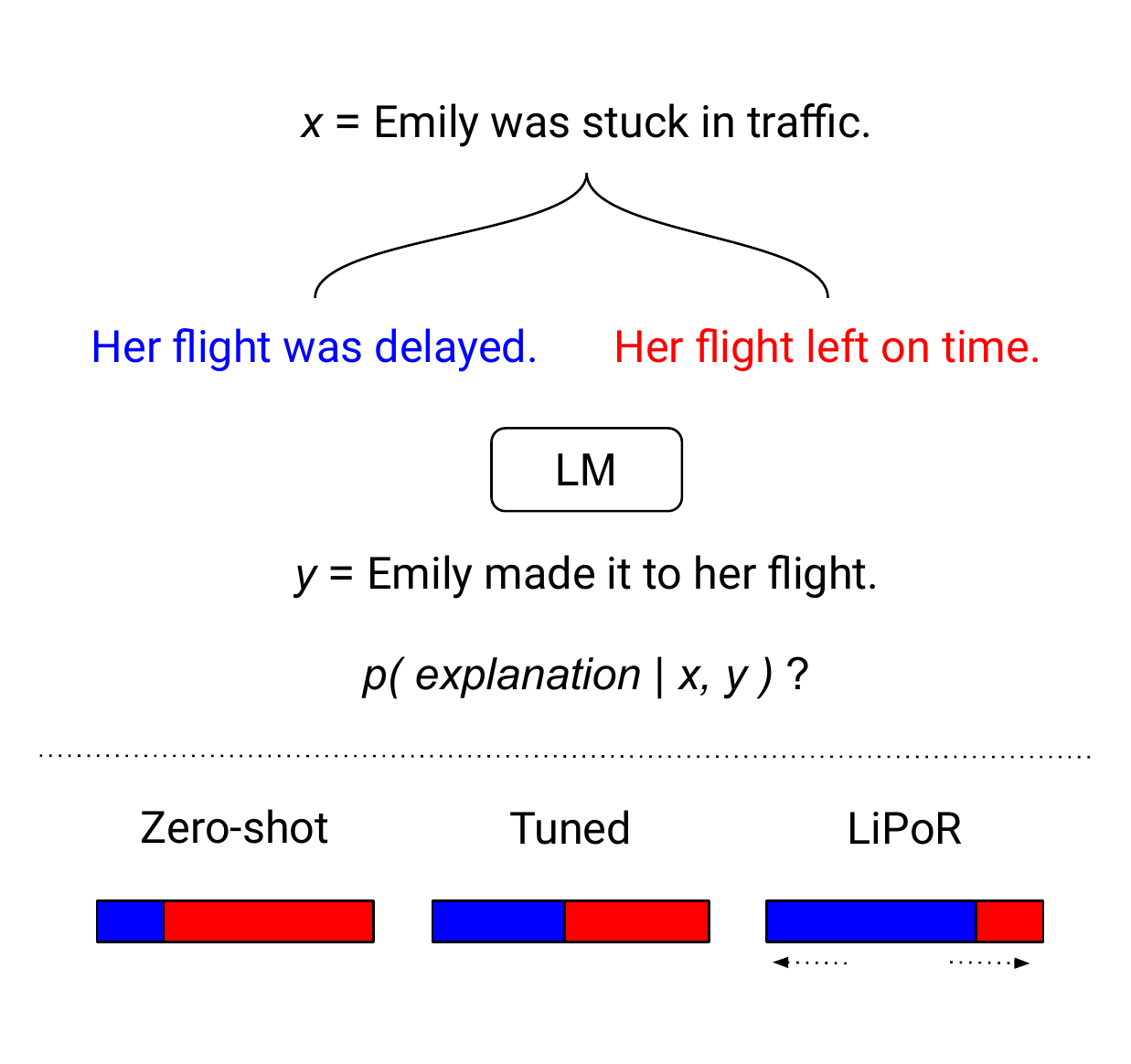}
\caption{
\textbf{Top:} An abductive reasoning example consisting of a context $x$, an outcome $y$, and two candidate explanations. The goal is to identify the plausible explanation given $x$ and $y$.
To predict an explanation, one can apply a pretrained language model (shown as LM) to score $y$ given $x$ and an explanation, and then compute the posterior probability for the explanation.
\textbf{Bottom:} Using a LM without fine-tuning (Zero-shot) leads to poor performance, whereas
a LM fine-tuned via max-marginal likelihood (Tuned) fails to distinguish the two explanations.
LiPoR is trained to partition the explanations in a mutually exclusive manner.
}
\label{fig:example}
\end{figure}

Abductive reasoning aims to find \emph{plausible} explanations for an event~\cite{paul1993approaches}.
Unlike deduction, which draws a firm conclusion from a set of premises, abduction requires reasoning from an outcome to plausible explanations.
Fig.~\ref{fig:example} (top) demonstrates the distinction: given only the context $x$, both the blue and the red sentences describe possible subsequent events; however, upon seeing the outcome $y$ only one of the two is a plausible explanation (although there may be others).
Humans apply abduction in everyday situations~\cite{andersen1973abductive} such as reading-between-the-lines~\cite{charniak1990probabilistic} and analyzing causes and effects~\cite{thagard1997abductive,pearl2018book}.
Learning to perform abduction is thus an important step towards building human-like machines with commonsense knowledge.

Abductive reasoning has been extensively studied in the setting where annotations are available~\cite{storks2019recent}.
However, because determining whether an explanation is plausible is a subjective and noisy process, annotating plausibility of explanations can be problematic for commonsense reasoning problems.
\citet{zhang2020winowhy} show that, in a dataset verification step where five annotators are asked to determine whether a hand-written explanation is plausible, they disagree with each other on 62.34\% of 1365 explanations.
This subjectivity thus introduces annotator-specific bias as has been seen in related tasks~\cite{elazar2021back,geva2019we}.
The potential bias in plausibility annotation motivates the study of learning to perform abductive reasoning without plausibility annotations.
Thus, we consider the setting where the context $x$ and outcome $y$ are observed, and models must learn to identify plausible explanations out of a given set of candidate explanations, without direct supervision over plausibility.

Rule-based methods use formal logic to reason about explanations~\cite{paul1993approaches}; however, their limited coverage prevents them from scaling to the full complexity of natural language.
Recently, pretrained language models, which have achieved remarkable 
performance on a range of NLP tasks~\cite{li2020unsupervised,wei2022finetuned}, hold the potential for zero-shot abductive reasoning. 
Specifically, \citet{bhagavatula2019abductive} directly estimate the probability of an explanation for an outcome through Bayes' Rule (\textit{Zero-shot} in Fig.~\ref{fig:example}).
In practice, however, this direct approach can often lead to performance that is only slightly better than random guessing~\cite{zhang2020winowhy,zhou-etal-2021-probing-commonsense}. 


To avoid these issues, we reduce abductive reasoning down to a single constraint --- an explanation must be \textit{plausible} or \textit{implausible}. This restriction, argued by \citet{gordon_hobbs_2017}, enforces that explanations are mutually exclusive; that is, one explanation being plausible automatically rules out some other explanations.
We introduce \textbf{Li}kelihood learning with \textbf{Po}sterior \textbf{R}egularization (LiPoR), an approach to perform abductive reasoning that only leverages mutual exclusivity of explanations and does not rely on plausibility annotations.
Specifically, we maximize the marginal likelihood of the outcome given the context and a set of explanations (\textit{Tuned} in Fig~\ref{fig:example}), then use posterior regularization to enforce mutual exclusion between plausible and implausible explanations (\textit{LiPoR} in Fig~\ref{fig:example}). We show how to impose this relation with a simple distributional constraint on the posterior of the model.
We empirically evaluate LiPoR on a diverse set of abductive reasoning datasets. Specifically, we consider four datasets under the abductive reasoning framework: $\alpha$NLI~\cite{bhagavatula2019abductive}, Sen-Making~\cite{wang2019does}, $\delta$-NLI~\cite{rudinger-etal-2020-thinking}, and WinoWhy~\cite{zhang2020winowhy}.
Results show that LiPoR consistently outperforms pretrained language models directly applied in a zero-shot manner and is comparable to different variants of a state-of-the-art knowledge-augmented zero-shot method~\cite{ma2021knowledge}.
As human-written explanation candidates are not always available during fine-tuning, we further evaluate LiPoR on the explanation candidates generated via prompting~\cite{brown2020language}.
We show that, even though automatically generated explanations are noisy, LiPoR can still leverage them and outperform strong zero-shot models including GPT3.


\section{Related Work}
\paragraph{Zero-shot commonsense reasoning.}
We categorize zero-shot approaches for commonsense reasoning into two groups.
The first group uses pretrained language models as a source of world knowledge.
\citet{shwartz2020unsupervised,zhou-etal-2021-think} query the language models with information seeking questions to identify background knowledge relevant to specific examples, and the answers returned by the models are later used as additional information for producing the final outputs.
\citet{dou2022improving} convert multiple-choice QA to cloze-style sentences and have the language models score different answers.
\citet{qin2020back} proposed a decoding algorithm that generates free-form explanations by considering the future contexts through back-propagation.
Our approach also uses pretrained language models as a source of knowledge, but we perform additional maximum likelihood fine-tuning to fit the abductive task data.

The second group leverages external knowledge bases (KBs).
\citet{bosselut2021dynamic} leverage COMET~\cite{bosselut2019comet}, a dynamic knowledge graph, to generate a chain of commonsense inferences based on contexts of QA examples, which can be treated as explanations.
\citet{banerjee2020self,ma2021knowledge} pretrain language models on artificial question answering (QA) datasets, created from knowledge graphs; a system trained on such datasets can directly perform zero-shot QA.
\citet{huang2021improving} formulate multiple-choice QA as natural language inference (NLI) and leverage both existing NLI datasets and KBs to identify answer choices in a zero-shot manner.

\paragraph{Relation to deductive reasoning.}
Both abduction and deduction have intermediate explanations.
Abductive reasoning infers the most likely explanation from outcomes.
In contrast, deductive reasoning infers a conclusion given a complete set of premises.
However, outcomes are often not a direct result of premises but come from a chain of reasoning over intermediate explanations. Identifying and providing the correct chain of reasoning is crucial to building trustworthy systems.

Within the realm of deduction there are several different approaches that utilize neural models. \citet{bostrom2021flexible} develop a pipeline to automatically construct training examples from Wikipedia, so that a system trained on such data is able to generate deductive inferences from natural language inputs without direct human supervision.
\citet{arabshahi2021conversational} present a neuro-symbolic theorem prover that extracts intermediate reasoning steps for understanding conversations.
\citet{rajani2019explain,tafjord2021proofwriter,nye2022show,wei2022chain} collect human annotated explanations for training interpretable systems which first generate intermediate explanations and then produce the final task outputs.

\paragraph{Explanations as latent variables.}
%
Modeling intermediate explanations as latent variables is a common approach, although training and inference details differ. Here we consider representative works in NLP.  \citet{zhou2020towards} apply a latent variable model to language understanding and train the model with variational expectation maximization.
Their method can generate free-form explanations but requires a small set of labeled examples for supervision.
\citet{zhou-etal-2021-probing-commonsense} apply such a model to probe dialogue generation in a zero-shot manner.
\citet{NEURIPS2020_92650b2e} apply a latent variable model to analyze gender bias in large pretrained language models by viewing the behaviors of neurons as unobserved explanations.
\citet{lei2016rationalizing,vafa2021rationales} apply such a model to identify rationales for sequence classification/generation, where rationales are a minimal subset of inputs or previous words that can lead to the same predictions.
LiPoR is a training scheme developed for learning such latent-variable models for abductive reasoning, which has a unique challenge of identifying multiple plausible explanations.

\section{Abductive Reasoning}
\begin{table}[t]
\centering
\resizebox{\linewidth}{!}{%
\begin{tabular}{p{0.24\linewidth}  p{0.01\linewidth}  p{0.73\linewidth}}
\toprule
\multirow{3}{*}{$\alpha$NLI} & \cellcolor{blue!15}$x$:           & \cellcolor{blue!15}it was a very hot summer day                                                                                                                                                      \\
                             & \cellcolor{pink!30}$z$: & \cellcolor{pink!30}\{he decided to run in the heat, \textbf{he drank a glass of ice cold water}\}                                                                                                             \\
                             & \cellcolor{green!15}$y$:           & \cellcolor{green!15} he felt much better                                                                                                                                                               \\ \hline
\multirow{2}{*}{Sen-Making}  & \cellcolor{pink!30}$z$: & \cellcolor{pink!30}\{\textbf{a restaurant does not have doctors or medical treatment}, a restaurant is usually too noisy for a patient, there are different types of restaurants in the city\}                \\
                             & \cellcolor{green!15}$y$:           & \cellcolor{green!15}it is not true that he was sent to a restaurant for treatment                                                                                                                       \\ \hline
\multirow{3}{*}{$\delta$-NLI} & \cellcolor{blue!15}$x$:           & \cellcolor{blue!15}four people and a child walking in the street                                                                                                                                     \\
                             & \cellcolor{pink!30}$z$: & \cellcolor{pink!30}\{\textbf{people from all over the world are gathered in the area}, \textbf{the people buy cotton candy from a booth}, the family is the only humans in the area, the family is walking their dog\} \\
                             & \cellcolor{green!15}$y$:           & \cellcolor{green!15}the family is enjoying the world's fair                                                                                                                                           \\ \hline
\multirow{3}{*}{WinoWhy}     & \cellcolor{blue!15}$x$:           & \cellcolor{blue!15}the fish ate the worm, it was hungry                                                                                                                                              \\
                             & \cellcolor{pink!30}$z$: & \cellcolor{pink!30}\{\textbf{hungry staff tend to eat}, worm is one being eaten, the worm is a common name for a variety of fish                                                                              \\
                             & \cellcolor{green!15}$y$:           & \cellcolor{green!15}therefore, it refers to the fish\\
\bottomrule
\end{tabular}%
}
\caption{Examples conversions from different datasets. Every dataset comes with candidate explanations (shown in the pink cells), and only a subset of them are plausible explanations (shown in boldface). We set $x$ in Sen-Making dataset to empty.}
\label{tab:conversion}
\end{table}


We consider four datasets that test abductive reasoning skills. While abduction can be difficult to pinpoint, we select datasets that obey the following criteria: 
there is a need for differentiating plausible explanations from implausible explanations, there is an observed outcome, and the outcome depends on intermediate explanations.
Based on these criteria, we use $\alpha$NLI~\cite{bhagavatula2019abductive}, Sen-Making~\cite{wang2019does}, $\delta$-NLI~\cite{rudinger-etal-2020-thinking}, and WinoWhy~\cite{zhang2020winowhy} as our target datasets.

To convert each to the abduction format, we first identify a context $x$, which sets a scope for candidate explanations $\mathcal{Z}$, as well as an outcome $y$.
The outcome could either be an event caused by $z$ or a conclusion reached by $z$.
Importantly, we differentiate explanation candidates $\mathcal{Z}$ as ones that are consistent with $x$, from plausible explanations that are consistent with both $x$ and $y$. A central assumption is that training abductive reasoning systems with the candidate set introduces less noise and subjectivity than directly supervising the systems with plausibility annotations.

Example conversions of each dataset are shown in Table~\ref{tab:conversion}.
Because $\alpha$NLI is designed as an abduction task, the conversion is straightforward.
Sen-Making is a benchmark that tests if a system can identify the reason why a statement is against common sense.
In this case, a context is not required.
We turn the nonsensical statement into a negative sentence, which becomes $y$.
Then the original answer choices become $z$.
$\delta$-NLI is a defeasible inference task, which requires deciding whether new evidence has strengthened or weakened the original hypothesis.
$\delta$-NLI is made of extensions to three existing inference datasets: SNLI~\cite{bowman2015large}, ATOMIC~\cite{sap2019atomic}, and SOCIAL-CHEM-101~\cite{forbes2020social}; each of them will be referred to as $\delta$-$N$ for brevity, where $N$ can be replaced by a dataset name.
We map premises and hypotheses to contexts and outcomes, respectively.
We then turn updates that strengthen a hypothesis into a plausible explanation and updates that weaken a hypothesis into an implausible explanation.
WinoWhy is a follow-up task for Winograd Schema Challenge (WSC)~\cite{levesque2012winograd}: Given the pronoun coreference resolution question and the answer from a WSC example, WinoWhy seeks to select all plausible reasons for why the pronoun is resolved to the answer.
We thus turn the question of the WSC example into a context $x$ and the answer into a declarative sentence $y$.

Notably these datasets differ in the number of plausible explanations, which we denote by a value $m\geq 1$. 
In $\alpha$NLI and Sen-Making, $m$ is fixed to 1 for all examples.
However, in $\delta$-NLI and WinoWhy, $m$ is variable, and we assume that half of explanations are plausible.
However these explanations are discrete; an explanation is either plausible or implausible.
A successful unsupervised system should assign high probabilities to plausible explanations and low probabilities to implausible explanations.
This discreteness is encoded into some of the tasks directly. For example,  \citet{bhagavatula2019abductive,zhang2020winowhy} instruct the annotators to make minimal possible changes to plausible explanations to produce implausible explanations, so that a system would fail if it predicts explanations based on superficial lexical features.

\section{LiPoR}

We now describe LiPoR, a method to adapt pretrained language models to incorporate mutual exclusivity between explanations. As we have seen, an abductive reasoning example consists of a context $x$, an observed outcome $y$, and an unobserved explanation $z\in\mathcal{Z}$, which, together with $x$, has led to $y$.
Importantly, the candidate set of explanations $\mathcal{Z}$ is given during training but the plausibility of each explanation is not.\footnote{While manually collecting $\mathcal{Z}$ can be expensive, we also show that $\mathcal{Z}$ can be also be obtained cheaply via language model prompting in Sec.~\ref{sec:analysis}.}
The goal of abductive reasoning is to produce a distribution over explanations $z$, defined by $p(z|x,y)$.
We are interested in modeling the joint distribution $p(y,z|x)$, which is factored as follows:
\begin{equation}
\label{eq:joint_prob}
    p(y,z|x)=p(y|x,z)p(z|x)
\end{equation}

Given Eq~\ref{eq:joint_prob}, the posterior distribution can be obtained via the Bayes' rule,
\begin{equation}
\label{bayes}
p(z|x,y)=\frac{p(y|z,x)p(z|x)}{p(y|x)}.
\end{equation}
Because $x$ itself does not provide further information for $z$, we set $p(z|x)$ to be a uniform distribution.
Therefore, we only parameterize $p(y|x,z)$.

\subsection{Baseline: Fine-tuning via Max-marginal Likelihood}
We note that any off-the-shelf pretrained language model can be applied to evaluate $p(z|x,y)$ for an abductive reasoning task in a zero-shot fashion.
To adapt the pretrained model to a specific task distribution without plausibility annotations, we maximize the following marginal likelihood function $\mathcal{L}(\cdot)$ with respect to parameters $\theta$ for all examples:
\begin{equation}
\label{eq:likelihood}
\mathcal{L}(\theta) = \text{log} \sum_{z\in\mathcal{Z}} p_{\theta}(y|x,z)p(z|x).
\end{equation}
Maximizing the marginal likelihood encourages the model to prefer explanations that assign the outcome high probability. Mechanically, the marginal likelihood requires computing the probability of the outcome given every explanation in the set $\mathcal{Z}$. Training then gives credit (gradient) to explanations that assign high probability to the outcome, encouraging the model to prefer explanations that explain the outcome.
We parameterize $p(y|x,z)$ by $\theta$, a language model, that takes ``$x \ [\text{SEP}] \ z$'' as input and returns a probability distribution over $y$.
By optimizing this objective, we find $\theta$ under which $p(y|x)$ has a high likelihood, thus shifting the pretrained model to the new task-specific distribution.
Furthermore, this objective does not require plausibility annotations for explanations.

\subsection{Incorporating Mutual Exclusivity}
\begin{figure}[t]
    \centering
    \includegraphics[width=0.8\linewidth]{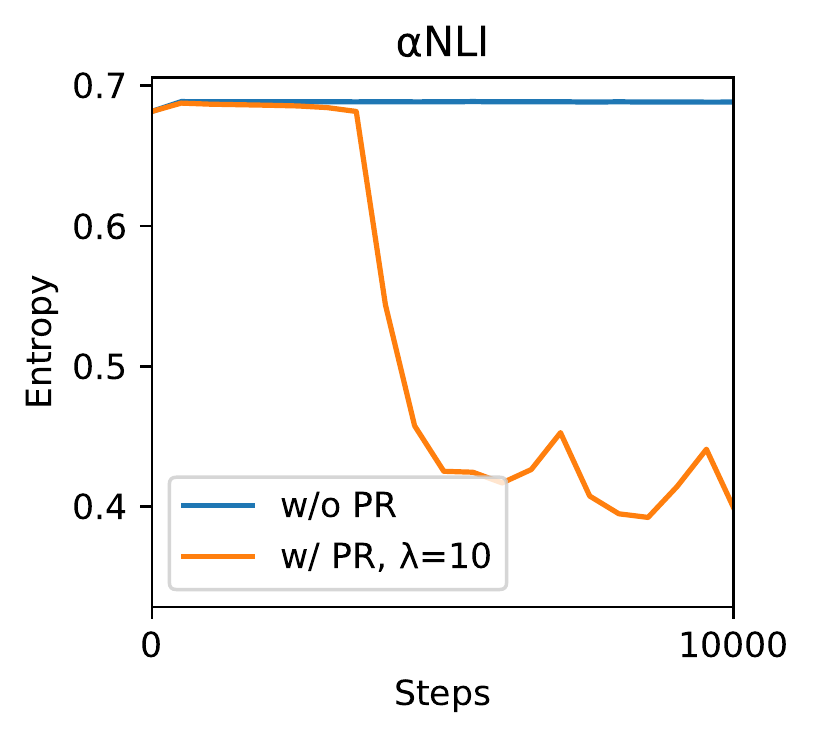}
    \caption{Entropy of $p(z|x,y)$ on $\alpha$NLI at different training steps. The orange line and the blue line represent with and without PR, respectively. Without PR the model never learns to distinguish between explanations.}
    \label{fig:pr_loss}
\end{figure}
The goal of abductive reasoning is to separate out plausible and implausible explanations. However, we note that $\mathcal{L}(\theta)$ itself only maximizes $p(y|x)$. In practice, this does not require the model to learn any distinctions 
between explanations, and we observe that in practice the approach learns to treat them all as plausible.
The blue line in Fig~\ref{fig:pr_loss} shows the entropy of $p(z|x,y)$ on the $\alpha$NLI dataset when fine-tuning a model with $\mathcal{L}(\theta)$.
We note that a uniform distribution of two categories has approximately an entropy of 0.6931, the upper bound on the entropy of $p(z|x,y)$ for the $\alpha$NLI examples.
Fine-tuning via max-marginal likelihood alone yields an entropy close to the upper bound, meaning the model believes that different $z$ explain $y$ equally well.

To impose the mutual exclusivity among explanations, we apply posterior regularization (PR), which places soft constraints on posterior distributions \citep{ganchev2010posterior}.
The posterior regularized likelihood shows as follows:
\begin{equation}
\label{eq:pr_likelihood}
\mathcal{L}_{PR}(\theta) = \mathcal{L}(\theta) - \lambda \Omega(p_{\theta}(z|x,y)).
\end{equation}
\begin{figure}
    \centering
    \includegraphics[width=.6\linewidth]{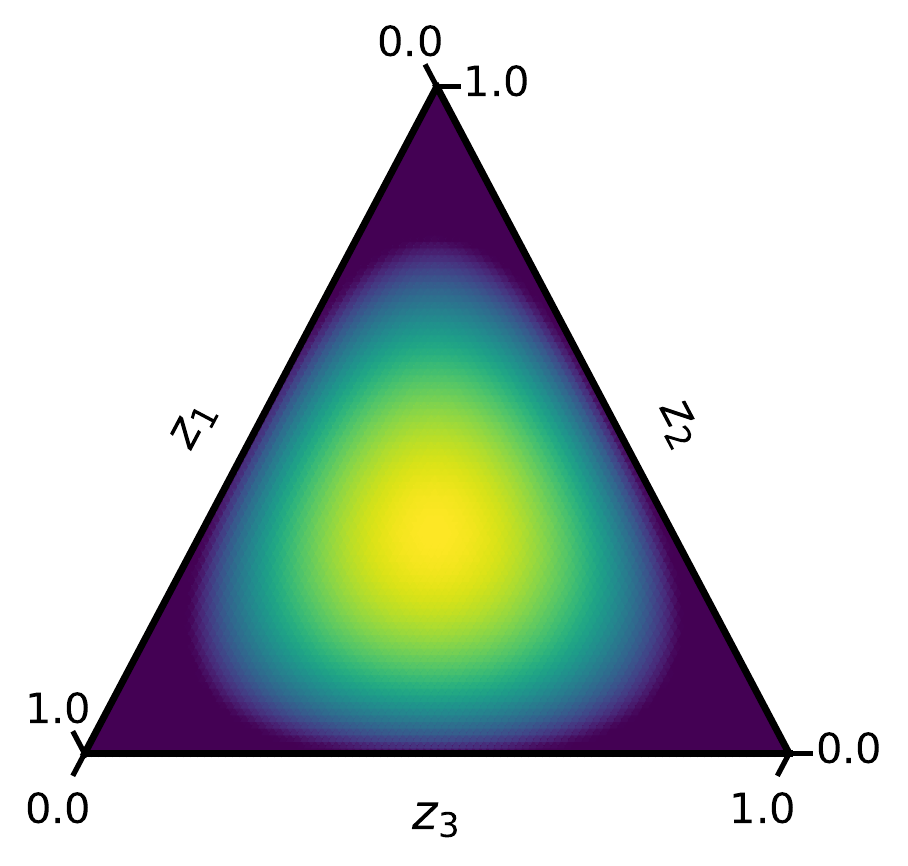}
    \caption{Visualization of $\Omega(\cdot)$ for $|\mathcal{Z}|=3$ and $m=2$. The lighter colors correspond to larger values. This constraint penalizes models that select too many plausible explanations.}
    \label{fig:ternary}
\end{figure}
To enforce a model to prefer specific explanations over the others, we choose $\Omega:
\mathcal{R}^{|\mathcal{Z}|} \rightarrow \mathcal{R}$ to be the following function, proposed in \citet{chen2020deep}:
\begin{equation}
\Omega(p(z|x,y)) = \maxB(H(p_\theta(z|x,y)), \text{ln}(m))
\end{equation}
$H(\cdot)$ is the entropy function.
In Fig.~\ref{fig:ternary}, we plot $\Omega(\cdot)$ when $|\mathcal{Z}|=3$ and $m=2$, which shows that distributions with a non-zero probability for the third explanation have larger $\Omega$ values.
$\Omega(\cdot)$ thus penalizes a posterior distribution that has an entropy higher than $\text{ln}(m)$, which sets an upper bound at the entropy of a distribution whose probability mass collapses to $m$ categories.
When $m=1$, $\Omega(\cdot)$ reduces to
\begin{equation}
    \Omega(p(z|x,y)) = H(p_\theta(z|x,y)).
\end{equation}
The orange line in Fig.~\ref{fig:pr_loss} shows that incorporating $\Omega(\cdot)$ enables the model to differentiate between different explanations.
Notice that, except for $m=1$, there is no guarantee that $\Omega(\cdot)$ penalizes all distributions that have probability mass in more than $m$ categories, but we will empirically justify that $\Omega(\cdot)$ eliminates undesired posterior distributions.
\begin{table*}[!t]
\centering
\resizebox{\textwidth}{!}{%
\begin{tabular}{p{0.15\linewidth} p{0.21\linewidth} rrrrrr}
\toprule
          & & \multicolumn{1}{l}{$\alpha$NLI} & \multicolumn{1}{l}{Sen-Making} & \multicolumn{1}{l}{$\delta$-ATOMIC} & \multicolumn{1}{l}{$\delta$-SNLI} & \multicolumn{1}{l}{$\delta$-SOCIAL} & \multicolumn{1}{l}{WinoWhy} \\ \midrule
&Previous Best & 65.50 & 45.60  &- &- &- &56.37 \\
&ZS GPT-NEO   & 57.47                           & 29.80                          & 47.53                             & 45.38                               & 51.69                               & 59.13                       \\
\multirow{2}{*}{w/o annotations}&ZS GPT3   & 67.54                           & 43.00                          & 50.73                             & 49.69                               & 49.22                               & 50.99                       \\
& ZS BART &50.96&47.80&59.05&55.12&52.58&45.69 \\
&Tuned BART &57.40&63.50&67.49&64.76&53.88&55.32 \\
&LiPoR    & \textbf{71.56}                  & \textbf{65.50}                & \textbf{76.82}                   & \textbf{65.26}                      & \textbf{57.19}                      & \textbf{69.88}              \\ \midrule
w/ annotations & RoBERTa & 85.60                           & 93.10                          & 78.30                             & 81.60                               & 86.20                               & 75.04\\ \midrule
 & KDDC-ATOMIC (N) & 70.80 & 51.00 &75.90 & 69.83 & 64.49 &42.44 \\
& KDDC-CWWV (N) & 70.00 & 45.70 &62.48 &63.24 &62.90 &40.45 \\
w/ KB & KDDC-CSKG (N) & 70.50 &49.60 &72.20 &69.93 &63.80 & 44.05 \\ 
& QNLI-ATOMIC (N) & - &- &- &- &- & 73.47 \\
 & Previous Best (Y)  & 87.30 &95.00&-&-&-& 87.55\\
\bottomrule
\end{tabular}%
}
\caption{
Accuracy for identifying plausible explanations using methods with and without plausibility annotations.
On each dataset, we boldface the best result within the methods without annotations.
Suffix (Y) / (N) denotes whether a knowledge-augmented method use (Y) or not use (N) annotations, respectively.
}
\label{tab:results}
\end{table*}

\section{Experimental Setup}
\label{sec:setup}

\paragraph{Metrics.} Accuracy is used to evaluate a system's predictive power.
For datasets with $m=1$, accuracy is computed with regards to each example (i.e., whether the plausible explanation has been identified for each example).
Otherwise, to stay consistent with evaluation in prior works, we compute accuracy with regards to each explanation (i.e., whether the plausibility of each explanation is correctly predicted).
Therefore, more weight will be given to the instances that have larger $|\mathcal{Z}|$ (within a single dataset, the variance of $|\mathcal{Z}|$ for different examples is very small).

\paragraph{Baselines.} We consider three groups of baselines: (1) methods that do not rely on plausibility annotations (shown as w/o annotations), (2) pretrained language models fine-tuned with plausibility annotations (shown as w/ annotations), and (3) methods that incorporate external knowledge bases (shown as ``w/ KBs''). For (1), we first consider previous best published results achieved by a RoBERTa-large model for $\alpha$NLI~\cite{ma2021knowledge}, by a BERT model for Sen-Making~\cite{wang2019does}, and a GPT-small model for WinoWhy~\cite{zhang2020winowhy} (all abbreviated as Prev. Best).
Additionally, we use GPT-Neo~\cite{gpt-neo}, GPT3 (text-davinci-002)~\cite{brown2020language}, and the BART-large model~\cite{lewis2020bart} to directly score $x \ [\text{SEP}] \ z \ [\text{SEP}] \ y$ for each $z$ in a zero-shot (ZS) manner.
We threshold the outputs of these models in the same way as done in our method to choose the plausible explanations.
Finally, we consider BART fine-tuned with Eq.~\ref{eq:likelihood} (Tuned BART) as a baseline to better understand the role of posterior regularization.
For (2), a RoBERTa-large model~\cite{liu2019roberta} is fine-tuned with plausibility annotations (abbreviated as RoBERTa).
For this baseline, we refer to the best result in the literature: \citet{ma2021knowledge} for $\alpha$NLI, \citet{wang-etal-2020-semeval} for Sen-Making, \citet{rudinger-etal-2020-thinking} for $\delta$-NLI, and \citet{zhang2020winowhy} for WinoWhy.
For (3), we run different variants of Knowledge-driven Data Construction (abbreviated as KDDC)~\cite{ma2021knowledge}, a method that leverages external knolwedge but not plausibility annotations.
We note that KDDC is designed to predict a single correct answer with argmax.
To handle the datasets that have more than one correct answers, we modify KDDC to choose the answers that have scores higher than the median.
We also include Knowledge-Enabled Natural Language Inference~\cite{huang2021improving} that is first supervised on QNLI~\cite{wang2018glue} and then incorporate ATOMIC at inference time for WinoWhy (abbreviated as QNLI-ATOMIC).
For models that use both external knowledge and plausibility annotations, we take RAINBOW~\cite{raina2005robust} for $\alpha$NLI, ECNU-SenseMaker~\cite{zhao-etal-2020-ecnu} for Sen-Making, and RoBERTa-Grande~\cite{zhang2020winowhy} for WinoWhy.

\paragraph{Implementation \& Hyperparameters.} We choose a BART-large model~\cite{lewis2020bart} to be $\theta$.
We train the model with the Hugging Face Transformers framework~\cite{wolf-etal-2020-transformers}.
We perform grid search with learning rates \{1e-6, 3e-6, 5e-6, 1e-5\}, batch sizes \{2,4,8,16\}, and $\lambda$ \{1e-2,1e-1,1,10\}.
We train 50 epochs for WinoWhy and 10 epochs for all other datasets.
We perform evaluation on dev sets every 500 steps.
We choose the checkpoint whose posterior distributions have the lowest average entropy on dev sets to run tests if the entropy starts to diverge during training.
If the entropy converges, we choose the checkpoint at the end of training.

Because there are not train/dev/test sets for WinoWhy, to perform a direct comparison with other methods, we do not split the dataset ourselves and simply train models on all of the data and choose the checkpoint based on loss values.

\paragraph{Automatic Candidate Generation}
\begin{figure}[t]
    \framebox{
    \parbox{0.45\textwidth}{
    \small
    {\color{blue} Prompt for plausible explanations}: Provide a brief explanation for why it is not sensible that $y$\newline
    {\color{blue}Prompt for implausible explanations}:
    Provide a brief explanation for why $y$\smallskip\newline
    $y$: He poured orange juice on his cereal. \smallskip\newline
    \textbf{In}: \textit{Provide a brief explanation for why it is not sensible that he poured orange juice on his cereal.} \newline
    \textbf{Out}: {\noindent \color{red} It is not sensible because orange juice does not go well with cereal.}\smallskip\newline
    \textbf{In}: \textit{Provide a brief explanation for why he poured orange juice on his cereal}\newline
    \textbf{Out}: {\noindent \color{red}He wanted to eat a healthy breakfast.}
    }
    }
    \caption{Prompts for producing competing explanations, followed by an example generation.} \label{fig:prompts}
\end{figure}
LiPoR assumes access to a candidate explanation set $\cal Z$ during training with human-written explanations. However, we may also want to use the model in domains without a candidate set.
We consider a variant that uses a noisy automatic candidate generation process. In this setting, set $\cal \tilde Z$ will contain a set of explanations with no guarantee that any are plausible.

To generate $\cal \tilde Z$ we utilize language model prompting with GPT3 (text-davinci-002) \cite{brown2020language}. Using prompt templates inspired by the instructions given to human annotators, we have the model generate explanations.
We show example prompts for the Sen-Making dataset in Fig.~\ref{fig:prompts}.
For datasets with fewer than 1000 unique contexts $x$ (i.e., $\delta$-NLI and Winowhy), we generate one plausible explanation and one implausible explanation for every $x$.
For the other datasets, we randomly sample 1000 unique contexts and otherwise stay the same.
We release the prompts as well as the generated explanations for every dataset in the supplementary materials.

In this setting, LiPoR uses a lower PR penalty $\lambda=0.1$.
We additionally consider two more baselines.
First, we score the most plausible explanation with the prompt as a prefix (denoted as Prompted GPT3).
Secondly, we supervise RoBERTa-large with the generated explanations.
\section{Results}
\label{sec:results}
\begin{table*}[t]
\small
\centering
\begin{tabular}{lrrrrrr}
\toprule
& \multicolumn{1}{l}{$\alpha$NLI} & \multicolumn{1}{l}{Sen-Making} & \multicolumn{1}{l}{$\delta$-ATOMIC} & \multicolumn{1}{l}{$\delta$-SNLI} & \multicolumn{1}{l}{$\delta$-SOCIAL} & \multicolumn{1}{l}{Winowhy} \\ \midrule
ZS GPT3 & \textbf{67.54} & 43.00 & 50.73 & 49.69 & 49.22 & 50.99 \\
Prompted GPT3 & 49.19 & 53.80 & 48.23 & 51.26 & 50.86 & 58.10 \\
LiPoR & 57.50 & \textbf{61.50} & \textbf{67.60} & \textbf{64.40} & \textbf{55.40} & \textbf{58.67} \\
RoBERTa (Y) & 53.71 & 61.30 & 62.74 & 57.81 & 51.78 & 42.13 \\
\bottomrule
\end{tabular}
\caption{Comparing LiPoR to several baselines on automatically generated explanation candidate sets. (Y) indicates that a method uses plausibility annotations.}
\label{tab:prompting_sen}
\end{table*}
\begin{table}[t]
\centering
\resizebox{0.45\linewidth}{!}{%
\begin{tabular}{lrr}
\toprule
                 & \multicolumn{1}{l}{{Tuned \color{green}\cmark}} & \multicolumn{1}{l}{Tuned {\color{red}\xmark}} \\ \midrule
ZS {\color{green}\cmark}   & 1140      & 419      \\
ZS {\color{red}\xmark} & 618      & 882    \\
\bottomrule
\end{tabular}%
} \hfill
\resizebox{0.51\linewidth}{!}{%
\begin{tabular}{lrr}
\toprule
       & \multicolumn{1}{l}{{LiPoR \color{green}\cmark}} & \multicolumn{1}{l}{LiPoR {\color{red}\xmark}} \\ \midrule
Tuned {\color{green}\cmark} & 1449                        & 309                        \\
Tuned {\color{red}\xmark} & 767                       & 534   \\
\bottomrule
\end{tabular}%
}
\caption{
\textbf{Left:} Comparison between ZS BART and Tuned BART on $\alpha$NLI. \textbf{Right:} Comparison between Tuned BART and LiPoR. {\{*\} \color{green}\cmark} and {\{*\} {\color{red}\xmark}} denote the number of instances for which plausible explanations are correctly / incorrectly identified by {\{*\}}, respectively.
}
\label{tab:confusion}
\end{table}
We summarize the results in Table~\ref{tab:results}.
First of all, LiPoR produces the best results compared to all other methods without plausibility annotations, including GPT3 which has many more parameters and is pretrained on more data.
We note that LiPoR consistently outperforms Tuned BART, suggesting that posterior regularization plays a positive role in selecting plausible explanations.
Compared to knowledge-augmented methods without plausibility annotations, LiPoR is able to produce better results on $\alpha$NLI, Sen-Making, and $\delta$-ATOMIC.
We note that $\delta$-NLI is in part created from knowledge bases, and therefore KDDC-* is particularly good at $\delta$-ATOMIC, $\delta$-SNLI, and $\delta$-SOCIAL, but fail on WinoWhy and Sen-Making.
Additionally, QNLI-ATOMIC outperforms LiPoR by 4 points on Winowhy, but this improvement is expected given how much related task data it was pretrained on.
Finally, LiPoR still cannot match the performance of RoBERTa trained with plausibility annotations.

In Table~\ref{tab:confusion}, we show the confusion matrices for comparing among ZS BART, Tuned BART, and LiPoR on the $\alpha$NLI test set.
Tuned BART and LiPoR make the same predictions on a majority of examples, and on the instances they disagree, LiPoR is able to correctly identify plausible explanations on twice as many examples.
We also observe a similar trend for ZS BART and Tuned BART.

\paragraph{Fine-tuning with Generated Explanations}
Table~\ref{tab:prompting_sen} compares LiPoR fine-tuned with generated explanation candidates to the best performing methods without plausibility annotations.
Even with noisy candidate sets, LiPoR is still able to leverage such data.
It outperforms zero-shot GPT3 methods and improves over Prompted GPT3.
Additionally, LiPoR is more robust than RoBERTa trained with plausibility annotations when such annotations are noisy.
Therefore, even though the generated explanations by themselves correlate weakly with plausibility, they can be used in LiPoR.
\section{Analysis}
\label{sec:analysis}
\paragraph{Preserving Plausible Candidates}
\begin{table}[t!]
\centering
\resizebox{\linewidth}{!}{%
\begin{tabular}{p{0.02\linewidth}  p{0.8\linewidth} p{0.07\linewidth} p{0.07\linewidth}}
\toprule
& Example & Y & N \\
\cellcolor{blue!15}$x$:                  & \cellcolor{blue!15}Sally went to Italy in the spring.                       & \cellcolor{blue!15}     & \cellcolor{blue!15}     \\
\cellcolor{pink!30} & \cellcolor{pink!30}Sally took a lot of pictures when she went sightseeing.                                      & \cellcolor{pink!30} 71.7 & \cellcolor{pink!30} 50.0 \\
\multirow{-2}{*}{\cellcolor{pink!30}$z$:}                   & \cellcolor{pink!30} Sally took pictures at every place she visited. & \cellcolor{pink!30} 28.3 & \cellcolor{pink!30} 50.0 \\
\cellcolor{green!15}$y$:                  & \cellcolor{green!15}When she got home, Sally showed her pictures to all her friends.        & \cellcolor{green!15}     & \cellcolor{green!15}\\ 
\hdashline
\cellcolor{blue!15}$x$:                  & \cellcolor{blue!15}Mike didn't study for a test.                       & \cellcolor{blue!15}     & \cellcolor{blue!15}     \\
\cellcolor{pink!30} & \cellcolor{pink!30} Mike was normally a good student.                                      & \cellcolor{pink!30} 100 & \cellcolor{pink!30} 50.0 \\
\multirow{-2}{*}{\cellcolor{pink!30}$z$:}                   & \cellcolor{pink!30} Everyone in class failed the test except for Mike. & \cellcolor{pink!30} 0 & \cellcolor{pink!30} 50.0 \\
\cellcolor{green!15}$y$:                  & \cellcolor{green!15}The teacher was very disappointed.        & \cellcolor{green!15}     & \cellcolor{green!15}\\ \hline
\cellcolor{yellow!30} {\noindent \color{red} \textbf{?}} &\multicolumn{3}{p{1.05\linewidth}}{\cellcolor{yellow!30} LiPoR assigns close probabilities to the indistinguishably likely explanations, while the supervised model collapses to one of the explanations.} \\
\bottomrule
\end{tabular}%
}
\caption{Comparison between posterior probabilities for each explanation produced by a RoBERTa model trained with plausibility annotations (Y) and LiPoR (N) on individual test examples, respectively.
}
\vspace*{-15pt}
\label{tab:bias}
\end{table}
Models trained to prefer single plausible explanations can become overconfident in their predictions. A major benefit of LiPoR is that it considers multiple plausible candidates. 
While LiPoR is fine-tuned to favor mutual exclusivity, we find that at test time it remains able to score multiple plausible explanations highly. 
Table~\ref{tab:bias} presents two examples in which both explanations are plausible.
The RoBERTa model trained with plausibility annotations produces posterior distributions that collapse to one explanation. However, LiPoR can assign significant probability to both explanations.


\paragraph{Qualitative Comparison}

\begin{table}[t!]
\centering
\resizebox{\linewidth}{!}{%
\begin{tabular}{p{0.02\linewidth}  p{0.8\linewidth} p{0.08\linewidth} p{0.07\linewidth}}
\toprule
& Example & -PR & +PR \\
\cellcolor{blue!15}$x$:                  & \cellcolor{blue!15}I love taking long warm \emph{showers}.                        & \cellcolor{blue!15}     & \cellcolor{blue!15}     \\
\cellcolor{pink!30}   & \cellcolor{pink!30}\emph{Showers} make me sleepy.                           & \cellcolor{pink!30} 50.3 & \cellcolor{pink!30} 6.0 \\
\multirow{-2}{*}{\cellcolor{pink!30}$z$:} & \cellcolor{pink!30}\textbf{Doing what you love is important.}    & \cellcolor{pink!30} 49.7 & \cellcolor{pink!30} 94.0 \\
\cellcolor{green!15}$y$:                  & \cellcolor{green!15}That's why I take two of them every day.                                             & \cellcolor{green!15}     & \cellcolor{green!15}      \\ \hdashline
\cellcolor{blue!15}$x$:                  & \cellcolor{blue!15}Neil wanted to see the \emph{mountains} of Asia.                        & \cellcolor{blue!15}     & \cellcolor{blue!15}     \\
\cellcolor{pink!30}   & \cellcolor{pink!30} \textbf{Neil booked a tripped online.}                           & \cellcolor{pink!30} 47.5 & \cellcolor{pink!30} 64.0 \\
\multirow{-2}{*}{\cellcolor{pink!30}$z$:} & \cellcolor{pink!30}Neil took a trip to see the Rocky \emph{mountains} instead.    & \cellcolor{pink!30} 52.5 & \cellcolor{pink!30} 36.0 \\
\cellcolor{green!15}$y$:                  & \cellcolor{green!15}Neil loved being so close to the \emph{mountains} in Nepal!                                             & \cellcolor{green!15}     & \cellcolor{green!15}      \\ \hline
\cellcolor{yellow!30} \faThumbsOUp &\multicolumn{3}{p{1.06\linewidth}}{\cellcolor{yellow!30} Fine-tuning (-PR) looks at superficial word co-occurrences, but LiPoR (+PR) tries to understand the true context.} \\ \bottomrule \\
\toprule
& Example & -PR & +PR \\
\cellcolor{blue!15}$x$:                  & \cellcolor{blue!15}We went to the park today.                       & \cellcolor{blue!15}     & \cellcolor{blue!15}     \\
\cellcolor{pink!30} & \cellcolor{pink!30}We were rained on!                                      & \cellcolor{pink!30} 53.5 & \cellcolor{pink!30} 3.9 \\
\multirow{-2}{*}{\cellcolor{pink!30}$z$:}                   & \cellcolor{pink!30} \textbf{We met a golden retriever puppy and he played with us.} & \cellcolor{pink!30} 46.5 & \cellcolor{pink!30} 96.1 \\
\cellcolor{green!15}$y$:                  & \cellcolor{green!15}I love going to the park!        & \cellcolor{green!15}     & \cellcolor{green!15}\\ \hdashline
\cellcolor{blue!15}$x$:                  & \cellcolor{blue!15}Before my lunch time I got a phone call.                       & \cellcolor{blue!15}     & \cellcolor{blue!15}     \\
\cellcolor{pink!30} & \cellcolor{pink!30}My best friend wanted to go on a trip.                                      & \cellcolor{pink!30} 50.5 & \cellcolor{pink!30} 40.9 \\
\multirow{-2}{*}{\cellcolor{pink!30}$z$:}                   & \cellcolor{pink!30} \textbf{My best friend wanted to try a new restaurant for lunch.} & \cellcolor{pink!30} 49.5 & \cellcolor{pink!30} 59.1 \\
\cellcolor{green!15}$y$:                  & \cellcolor{green!15}We had an amazing time!        & \cellcolor{green!15}     & \cellcolor{green!15}\\ \hline
\cellcolor{yellow!30} \faThumbsOUp &\multicolumn{3}{p{1.06\linewidth}}{\cellcolor{yellow!30} LiPoR (+PR) is able to correct the bias towards shorter explanations.} \\
\bottomrule  \\
\end{tabular}%
}
\caption{Comparison between posterior probabilities for each explanation produced by fine-tuning (-PR) and LiPoR (+PR) on individual test examples, respectively. The two tables consist of examples where LiPoR successfully corrects the mistakes made by fine-tuning. The plausible explanation labeled by human annotators are in boldface.}
\label{tab:ll-pr}
\vspace*{-10pt}
\end{table}

Table~\ref{tab:ll-pr} presents a number of examples accompanied with the predictions made by fine-tuning via max-marginal likelihood (-PR) and LiPoR (+PR) side by side.
The two examples on the top are among the more difficult abduction examples: the first example requires a model to draw a connection between abstract concepts and concrete objects (``what you love'' $\rightarrow$ ``taking long warm showers''); the second example requires a model to figure out an inclusion relation (Nepal is a country in Asia).
We italicize the words that co-occur across $x,z$ and $y$, 
and we speculate that fine-tuning chooses the wrong explanations because of lexical overlap shortcuts.
LiPoR, however, was able to correctly flip these predictions with high confidence.

The two examples on the bottom are those for which Tuned BART fails to identify the plausible explanation because one explanation is short and the other is long.
Again, LiPoR is able to correct these mistakes.
Furthermore, the probability produced by LiPoR for each explanation also reflects the model's confidence to a certain degree.
In the first example, ``we met a golden retriever puppy and he played with us'' is a much better explanation than ``we were rained on,'' because one does not need to go to a park to experience rain.
As a result, the difference between probabilities for the two explanations is 92.2\%.
For the second example, ``we had an amazing time'' could refer to both trying out a new restaurant and going on a trip.
The phone call was received before lunch time makes the second explanation more likely, but the first explanation can still be what actually happened.
As a result, LiPoR assigns 40.9\% to the ``trip'' explanation and 59.1\% to the ``restaurant'' explanation, leading to a smaller gap than that of the first example.

\section{Conclusion}
We introduce LiPoR, which fine-tunes pretrained language models on abductive reasoning tasks without plausibility annotations.
Results shows that LiPoR achieves comparable performance to that of knowledge-augmented zero-shot methods.


\section*{Ethical Statement}
LiPoR shares similar concerns with other contemporary approaches for performing commonsense reasoning.
Specifically, because LiPoR exploits the knowledge already present in pretrained language models, it can potentially reinforce existing harmful biases in such models.

\section*{Acknowledgement}

AR and JC are supported by a Sloan Fellowship, NSF CAREER \#2037519, and NSF \#1901030. CC and WZ are supported by NSF \#1815455.

\bibliography{anthology}

\begin{thebibliography}{50}
\expandafter\ifx\csname natexlab\endcsname\relax\def\natexlab#1{#1}\fi

\bibitem[{Andersen(1973)}]{andersen1973abductive}
Henning Andersen. 1973.
\newblock Abductive and deductive change.
\newblock \emph{Language}, pages 765--793.

\bibitem[{Arabshahi et~al.(2021)Arabshahi, Lee, Gawarecki, Mazaitis, Azaria,
  and Mitchell}]{arabshahi2021conversational}
Forough Arabshahi, Jennifer Lee, Mikayla Gawarecki, Kathryn Mazaitis, Amos
  Azaria, and Tom Mitchell. 2021.
\newblock Conversational neuro-symbolic commonsense reasoning.
\newblock In \emph{Proceedings of the AAAI Conference on Artificial
  Intelligence}, volume~35, pages 4902--4911.

\bibitem[{Banerjee and Baral(2020)}]{banerjee2020self}
Pratyay Banerjee and Chitta Baral. 2020.
\newblock Self-supervised knowledge triplet learning for zero-shot question
  answering.
\newblock In \emph{Proceedings of the 2020 Conference on Empirical Methods in
  Natural Language Processing (EMNLP)}, pages 151--162.

\bibitem[{Bhagavatula et~al.(2019)Bhagavatula, Le~Bras, Malaviya, Sakaguchi,
  Holtzman, Rashkin, Downey, Yih, and Choi}]{bhagavatula2019abductive}
Chandra Bhagavatula, Ronan Le~Bras, Chaitanya Malaviya, Keisuke Sakaguchi, Ari
  Holtzman, Hannah Rashkin, Doug Downey, Wen-tau Yih, and Yejin Choi. 2019.
\newblock Abductive commonsense reasoning.
\newblock In \emph{International Conference on Learning Representations}.

\bibitem[{Black et~al.(2021)Black, Leo, Wang, Leahy, and Biderman}]{gpt-neo}
Sid Black, Gao Leo, Phil Wang, Connor Leahy, and Stella Biderman. 2021.
\newblock \href {https://doi.org/10.5281/zenodo.5297715} {{GPT-Neo: Large Scale
  Autoregressive Language Modeling with Mesh-Tensorflow}}.
\newblock {If you use this software, please cite it using these metadata.}

\bibitem[{Bosselut et~al.(2021)Bosselut, Le~Bras, and
  Choi}]{bosselut2021dynamic}
Antoine Bosselut, Ronan Le~Bras, and Yejin Choi. 2021.
\newblock Dynamic neuro-symbolic knowledge graph construction for zero-shot
  commonsense question answering.
\newblock In \emph{Proceedings of the 35th AAAI Conference on Artificial
  Intelligence (AAAI)}.

\bibitem[{Bosselut et~al.(2019)Bosselut, Rashkin, Sap, Malaviya, Celikyilmaz,
  and Choi}]{bosselut2019comet}
Antoine Bosselut, Hannah Rashkin, Maarten Sap, Chaitanya Malaviya, Asli
  Celikyilmaz, and Yejin Choi. 2019.
\newblock Comet: Commonsense transformers for automatic knowledge graph
  construction.
\newblock In \emph{Proceedings of the 57th Annual Meeting of the Association
  for Computational Linguistics}, pages 4762--4779.

\bibitem[{Bostrom et~al.(2021)Bostrom, Zhao, Chaudhuri, and
  Durrett}]{bostrom2021flexible}
Kaj Bostrom, Xinyu Zhao, Swarat Chaudhuri, and Greg Durrett. 2021.
\newblock Flexible generation of natural language deductions.
\newblock In \emph{Proceedings of the 2021 Conference on Empirical Methods in
  Natural Language Processing}, pages 6266--6278.

\bibitem[{Bowman et~al.(2015)Bowman, Angeli, Potts, and
  Manning}]{bowman2015large}
Samuel Bowman, Gabor Angeli, Christopher Potts, and Christopher~D Manning.
  2015.
\newblock A large annotated corpus for learning natural language inference.
\newblock In \emph{Proceedings of the 2015 Conference on Empirical Methods in
  Natural Language Processing}, pages 632--642.

\bibitem[{Brown et~al.(2020)Brown, Mann, Ryder, Subbiah, Kaplan, Dhariwal,
  Neelakantan, Shyam, Sastry, Askell et~al.}]{brown2020language}
Tom Brown, Benjamin Mann, Nick Ryder, Melanie Subbiah, Jared~D Kaplan, Prafulla
  Dhariwal, Arvind Neelakantan, Pranav Shyam, Girish Sastry, Amanda Askell,
  et~al. 2020.
\newblock Language models are few-shot learners.
\newblock \emph{Advances in neural information processing systems},
  33:1877--1901.

\bibitem[{Charniak and Shimony(1990)}]{charniak1990probabilistic}
Eugene Charniak and Solomon~E Shimony. 1990.
\newblock Probabilistic semantics for cost based abduction.
\newblock In \emph{Proceedings of the eighth National conference on Artificial
  intelligence-Volume 1}, pages 106--111.

\bibitem[{Chen et~al.(2020)Chen, Bai, Zhao, Ament, Gregoire, and
  Gomes}]{chen2020deep}
Di~Chen, Yiwei Bai, Wenting Zhao, Sebastian Ament, John Gregoire, and Carla
  Gomes. 2020.
\newblock Deep reasoning networks for unsupervised pattern de-mixing with
  constraint reasoning.
\newblock In \emph{International Conference on Machine Learning}, pages
  1500--1509. PMLR.

\bibitem[{Dou and Peng(2022)}]{dou2022improving}
Zi-Yi Dou and Nanyun Peng. 2022.
\newblock Zero-shot commonsense question answering with cloze translation and
  consistency optimization.
\newblock In \emph{The Thirty-Sixth AAAI Conference on Artificial Intelligence
  (AAAI)}.

\bibitem[{Elazar et~al.(2021)Elazar, Zhang, Goldberg, and
  Roth}]{elazar2021back}
Yanai Elazar, Hongming Zhang, Yoav Goldberg, and Dan Roth. 2021.
\newblock Back to square one: Artifact detection, training and commonsense
  disentanglement in the winograd schema.
\newblock In \emph{Proceedings of the 2021 Conference on Empirical Methods in
  Natural Language Processing}, pages 10486--10500.

\bibitem[{Forbes et~al.(2020)Forbes, Hwang, Shwartz, Sap, and
  Choi}]{forbes2020social}
Maxwell Forbes, Jena~D Hwang, Vered Shwartz, Maarten Sap, and Yejin Choi. 2020.
\newblock Social chemistry 101: Learning to reason about social and moral
  norms.
\newblock In \emph{Proceedings of the 2020 Conference on Empirical Methods in
  Natural Language Processing (EMNLP)}, pages 653--670.

\bibitem[{Ganchev et~al.(2010)Ganchev, Gra{\c{c}}a, Gillenwater, and
  Taskar}]{ganchev2010posterior}
Kuzman Ganchev, Joao Gra{\c{c}}a, Jennifer Gillenwater, and Ben Taskar. 2010.
\newblock Posterior regularization for structured latent variable models.
\newblock \emph{The Journal of Machine Learning Research}, 11:2001--2049.

\bibitem[{Geva et~al.(2019)Geva, Goldberg, and Berant}]{geva2019we}
Mor Geva, Yoav Goldberg, and Jonathan Berant. 2019.
\newblock Are we modeling the task or the annotator? an investigation of
  annotator bias in natural language understanding datasets.
\newblock In \emph{Proceedings of the 2019 Conference on Empirical Methods in
  Natural Language Processing and the 9th International Joint Conference on
  Natural Language Processing (EMNLP-IJCNLP)}, pages 1161--1166.

\bibitem[{Gordon and Hobbs(2017)}]{gordon_hobbs_2017}
Andrew~S. Gordon and Jerry~R. Hobbs. 2017.
\newblock \href {https://doi.org/10.1017/9781316584705.026}
  {\emph{Explanation}}, page 299–305. Cambridge University Press.

\bibitem[{Huang et~al.(2021)Huang, He, and Liu}]{huang2021improving}
Canming Huang, Weinan He, and Yongmei Liu. 2021.
\newblock Improving unsupervised commonsense reasoning using knowledge-enabled
  natural language inference.
\newblock In \emph{Findings of the Association for Computational Linguistics:
  EMNLP 2021}, pages 4875--4885.

\bibitem[{Lei et~al.(2016)Lei, Barzilay, and Jaakkola}]{lei2016rationalizing}
Tao Lei, Regina Barzilay, and Tommi Jaakkola. 2016.
\newblock Rationalizing neural predictions.
\newblock In \emph{Proceedings of the 2016 Conference on Empirical Methods in
  Natural Language Processing}, pages 107--117.

\bibitem[{Levesque et~al.(2012)Levesque, Davis, and
  Morgenstern}]{levesque2012winograd}
Hector Levesque, Ernest Davis, and Leora Morgenstern. 2012.
\newblock The winograd schema challenge.
\newblock In \emph{Thirteenth international conference on the principles of
  knowledge representation and reasoning}.

\bibitem[{Lewis et~al.(2020)Lewis, Liu, Goyal, Ghazvininejad, Mohamed, Levy,
  Stoyanov, and Zettlemoyer}]{lewis2020bart}
Mike Lewis, Yinhan Liu, Naman Goyal, Marjan Ghazvininejad, Abdelrahman Mohamed,
  Omer Levy, Veselin Stoyanov, and Luke Zettlemoyer. 2020.
\newblock Bart: Denoising sequence-to-sequence pre-training for natural
  language generation, translation, and comprehension.
\newblock In \emph{Proceedings of the 58th Annual Meeting of the Association
  for Computational Linguistics}, pages 7871--7880.

\bibitem[{Li et~al.(2020)Li, Li, Mou, Jiang, Lyu, and
  King}]{li2020unsupervised}
Jingjing Li, Zichao Li, Lili Mou, Xin Jiang, Michael Lyu, and Irwin King. 2020.
\newblock Unsupervised text generation by learning from search.
\newblock \emph{Advances in Neural Information Processing Systems},
  33:10820--10831.

\bibitem[{Liu et~al.(2019)Liu, Ott, Goyal, Du, Joshi, Chen, Levy, Lewis,
  Zettlemoyer, and Stoyanov}]{liu2019roberta}
Yinhan Liu, Myle Ott, Naman Goyal, Jingfei Du, Mandar Joshi, Danqi Chen, Omer
  Levy, Mike Lewis, Luke Zettlemoyer, and Veselin Stoyanov. 2019.
\newblock Roberta: A robustly optimized bert pretraining approach.
\newblock \emph{arXiv preprint arXiv:1907.11692}.

\bibitem[{Ma et~al.(2021)Ma, Ilievski, Francis, Bisk, Nyberg, and
  Oltramari}]{ma2021knowledge}
Kaixin Ma, Filip Ilievski, Jonathan Francis, Yonatan Bisk, Eric Nyberg, and
  Alessandro Oltramari. 2021.
\newblock Knowledge-driven data construction for zero-shot evaluation in
  commonsense question answering.
\newblock In \emph{35th AAAI Conference on Artificial Intelligence}.

\bibitem[{Nye et~al.(2022)Nye, Andreassen, Gur-Ari, Michalewski, Austin,
  Bieber, Dohan, Lewkowycz, Bosma, Luan, Sutton, and Odena}]{nye2022show}
Maxwell Nye, Anders~Johan Andreassen, Guy Gur-Ari, Henryk Michalewski, Jacob
  Austin, David Bieber, David Dohan, Aitor Lewkowycz, Maarten Bosma, David
  Luan, Charles Sutton, and Augustus Odena. 2022.
\newblock \href {https://openreview.net/forum?id=HBlx2idbkbq} {Show your work:
  Scratchpads for intermediate computation with language models}.
\newblock In \emph{Deep Learning for Code Workshop}.

\bibitem[{Paul(1993)}]{paul1993approaches}
Gabriele Paul. 1993.
\newblock Approaches to abductive reasoning: an overview.
\newblock \emph{Artificial intelligence review}, 7(2):109--152.

\bibitem[{Pearl and Mackenzie(2018)}]{pearl2018book}
Judea Pearl and Dana Mackenzie. 2018.
\newblock \emph{The book of why: the new science of cause and effect}.
\newblock Basic books.

\bibitem[{Qin et~al.(2020)Qin, Shwartz, West, Bhagavatula, Hwang, Le~Bras,
  Bosselut, and Choi}]{qin2020back}
Lianhui Qin, Vered Shwartz, Peter West, Chandra Bhagavatula, Jena~D Hwang,
  Ronan Le~Bras, Antoine Bosselut, and Yejin Choi. 2020.
\newblock Back to the future: Unsupervised backprop-based decoding for
  counterfactual and abductive commonsense reasoning.
\newblock In \emph{Proceedings of the 2020 Conference on Empirical Methods in
  Natural Language Processing (EMNLP)}, pages 794--805.

\bibitem[{Raina et~al.(2005)Raina, Ng, and Manning}]{raina2005robust}
Rajat Raina, Andrew~Y Ng, and Christopher~D Manning. 2005.
\newblock Robust textual inference via learning and abductive reasoning.
\newblock In \emph{AAAI}, pages 1099--1105.

\bibitem[{Rajani et~al.(2019)Rajani, McCann, Xiong, and
  Socher}]{rajani2019explain}
Nazneen~Fatema Rajani, Bryan McCann, Caiming Xiong, and Richard Socher. 2019.
\newblock Explain yourself! leveraging language models for commonsense
  reasoning.
\newblock In \emph{Proceedings of the 57th Annual Meeting of the Association
  for Computational Linguistics}, pages 4932--4942.

\bibitem[{Rudinger et~al.(2020)Rudinger, Shwartz, Hwang, Bhagavatula, Forbes,
  Le~Bras, Smith, and Choi}]{rudinger-etal-2020-thinking}
Rachel Rudinger, Vered Shwartz, Jena~D. Hwang, Chandra Bhagavatula, Maxwell
  Forbes, Ronan Le~Bras, Noah~A. Smith, and Yejin Choi. 2020.
\newblock \href {https://doi.org/10.18653/v1/2020.findings-emnlp.418} {Thinking
  like a skeptic: Defeasible inference in natural language}.
\newblock In \emph{Findings of the Association for Computational Linguistics:
  EMNLP 2020}, pages 4661--4675, Online. Association for Computational
  Linguistics.

\bibitem[{Sap et~al.(2019)Sap, Le~Bras, Allaway, Bhagavatula, Lourie, Rashkin,
  Roof, Smith, and Choi}]{sap2019atomic}
Maarten Sap, Ronan Le~Bras, Emily Allaway, Chandra Bhagavatula, Nicholas
  Lourie, Hannah Rashkin, Brendan Roof, Noah~A Smith, and Yejin Choi. 2019.
\newblock Atomic: An atlas of machine commonsense for if-then reasoning.
\newblock In \emph{Proceedings of the AAAI Conference on Artificial
  Intelligence}, volume~33, pages 3027--3035.

\bibitem[{Shwartz et~al.(2020)Shwartz, West, Le~Bras, Bhagavatula, and
  Choi}]{shwartz2020unsupervised}
Vered Shwartz, Peter West, Ronan Le~Bras, Chandra Bhagavatula, and Yejin Choi.
  2020.
\newblock Unsupervised commonsense question answering with self-talk.
\newblock In \emph{Proceedings of the 2020 Conference on Empirical Methods in
  Natural Language Processing (EMNLP)}, pages 4615--4629.

\bibitem[{Storks et~al.(2019)Storks, Gao, and Chai}]{storks2019recent}
Shane Storks, Qiaozi Gao, and Joyce~Y Chai. 2019.
\newblock Recent advances in natural language inference: A survey of
  benchmarks, resources, and approaches.
\newblock \emph{arXiv preprint arXiv:1904.01172}.

\bibitem[{Tafjord et~al.(2021)Tafjord, Dalvi, and
  Clark}]{tafjord2021proofwriter}
Oyvind Tafjord, Bhavana Dalvi, and Peter Clark. 2021.
\newblock Proofwriter: Generating implications, proofs, and abductive
  statements over natural language.
\newblock In \emph{Findings of the Association for Computational Linguistics:
  ACL-IJCNLP 2021}, pages 3621--3634.

\bibitem[{Thagard and Shelley(1997)}]{thagard1997abductive}
Paul Thagard and Cameron Shelley. 1997.
\newblock Abductive reasoning: Logic, visual thinking, and coherence.
\newblock In \emph{Logic and scientific methods}, pages 413--427. Springer.

\bibitem[{Vafa et~al.(2021)Vafa, Deng, Blei, and Rush}]{vafa2021rationales}
Keyon Vafa, Yuntian Deng, David Blei, and Alexander~M Rush. 2021.
\newblock Rationales for sequential predictions.
\newblock In \emph{Proceedings of the 2021 Conference on Empirical Methods in
  Natural Language Processing}, pages 10314--10332.

\bibitem[{Vig et~al.(2020)Vig, Gehrmann, Belinkov, Qian, Nevo, Singer, and
  Shieber}]{NEURIPS2020_92650b2e}
Jesse Vig, Sebastian Gehrmann, Yonatan Belinkov, Sharon Qian, Daniel Nevo,
  Yaron Singer, and Stuart Shieber. 2020.
\newblock \href
  {https://proceedings.neurips.cc/paper/2020/file/92650b2e92217715fe312e6fa7b90d82-Paper.pdf}
  {Investigating gender bias in language models using causal mediation
  analysis}.
\newblock In \emph{Advances in Neural Information Processing Systems},
  volume~33, pages 12388--12401. Curran Associates, Inc.

\bibitem[{Wang et~al.(2018)Wang, Singh, Michael, Hill, Levy, and
  Bowman}]{wang2018glue}
Alex Wang, Amanpreet Singh, Julian Michael, Felix Hill, Omer Levy, and Samuel
  Bowman. 2018.
\newblock Glue: A multi-task benchmark and analysis platform for natural
  language understanding.
\newblock In \emph{Proceedings of the 2018 EMNLP Workshop BlackboxNLP:
  Analyzing and Interpreting Neural Networks for NLP}, pages 353--355.

\bibitem[{Wang et~al.(2020)Wang, Liang, Jin, Wang, Zhu, and
  Zhang}]{wang-etal-2020-semeval}
Cunxiang Wang, Shuailong Liang, Yili Jin, Yilong Wang, Xiaodan Zhu, and Yue
  Zhang. 2020.
\newblock {S}em{E}val-2020 task 4: Commonsense validation and explanation.
\newblock In \emph{Proceedings of The 14th International Workshop on Semantic
  Evaluation}. Association for Computational Linguistics.

\bibitem[{Wang et~al.(2019)Wang, Liang, Zhang, Li, and Gao}]{wang2019does}
Cunxiang Wang, Shuailong Liang, Yue Zhang, Xiaonan Li, and Tian Gao. 2019.
\newblock Does it make sense? and why? a pilot study for sense making and
  explanation.
\newblock In \emph{Proceedings of the 57th Annual Meeting of the Association
  for Computational Linguistics}, pages 4020--4026.

\bibitem[{Wei et~al.(2022{\natexlab{a}})Wei, Bosma, Zhao, Guu, Yu, Lester, Du,
  Dai, and Le}]{wei2022finetuned}
Jason Wei, Maarten Bosma, Vincent Zhao, Kelvin Guu, Adams~Wei Yu, Brian Lester,
  Nan Du, Andrew~M. Dai, and Quoc~V Le. 2022{\natexlab{a}}.
\newblock \href {https://openreview.net/forum?id=gEZrGCozdqR} {Finetuned
  language models are zero-shot learners}.
\newblock In \emph{International Conference on Learning Representations}.

\bibitem[{Wei et~al.(2022{\natexlab{b}})Wei, Wang, Schuurmans, Bosma, Chi, Le,
  and Zhou}]{wei2022chain}
Jason Wei, Xuezhi Wang, Dale Schuurmans, Maarten Bosma, Ed~Chi, Quoc Le, and
  Denny Zhou. 2022{\natexlab{b}}.
\newblock Chain of thought prompting elicits reasoning in large language
  models.
\newblock \emph{arXiv preprint arXiv:2201.11903}.

\bibitem[{Wolf et~al.(2020)Wolf, Debut, Sanh, Chaumond, Delangue, Moi, Cistac,
  Rault, Louf, Funtowicz, Davison, Shleifer, von Platen, Ma, Jernite, Plu, Xu,
  Scao, Gugger, Drame, Lhoest, and Rush}]{wolf-etal-2020-transformers}
Thomas Wolf, Lysandre Debut, Victor Sanh, Julien Chaumond, Clement Delangue,
  Anthony Moi, Pierric Cistac, Tim Rault, Rémi Louf, Morgan Funtowicz, Joe
  Davison, Sam Shleifer, Patrick von Platen, Clara Ma, Yacine Jernite, Julien
  Plu, Canwen Xu, Teven~Le Scao, Sylvain Gugger, Mariama Drame, Quentin Lhoest,
  and Alexander~M. Rush. 2020.
\newblock \href {https://www.aclweb.org/anthology/2020.emnlp-demos.6}
  {Transformers: State-of-the-art natural language processing}.
\newblock In \emph{Proceedings of the 2020 Conference on Empirical Methods in
  Natural Language Processing: System Demonstrations}, pages 38--45, Online.
  Association for Computational Linguistics.

\bibitem[{Zhang et~al.(2020)Zhang, Zhao, and Song}]{zhang2020winowhy}
Hongming Zhang, Xinran Zhao, and Yangqiu Song. 2020.
\newblock Winowhy: A deep diagnosis of essential commonsense knowledge for
  answering winograd schema challenge.
\newblock In \emph{Proceedings of the 58th Annual Meeting of the Association
  for Computational Linguistics}, pages 5736--5745.

\bibitem[{Zhao et~al.(2020)Zhao, Tao, Zhou, Wang, Lin, and
  He}]{zhao-etal-2020-ecnu}
Qian Zhao, Siyu Tao, Jie Zhou, Linlin Wang, Xin Lin, and Liang He. 2020.
\newblock \href {https://doi.org/10.18653/v1/2020.semeval-1.48}
  {{ECNU}-{S}ense{M}aker at {S}em{E}val-2020 task 4: Leveraging heterogeneous
  knowledge resources for commonsense validation and explanation}.
\newblock In \emph{Proceedings of the Fourteenth Workshop on Semantic
  Evaluation}, pages 401--410, Barcelona (online). International Committee for
  Computational Linguistics.

\bibitem[{Zhou et~al.(2021{\natexlab{a}})Zhou, Hedayatnia, Gopalakrishnan, Kim,
  Pujara, Ren, Liu, and Hakkani-Tur}]{zhou-etal-2021-think}
Pei Zhou, Behnam Hedayatnia, Karthik Gopalakrishnan, Seokhwan Kim, Jay Pujara,
  Xiang Ren, Yang Liu, and Dilek Hakkani-Tur. 2021{\natexlab{a}}.
\newblock \href {https://doi.org/10.18653/v1/2021.nlp4convai-1.23} {Think
  before you speak: Learning to generate implicit knowledge for response
  generation by self-talk}.
\newblock In \emph{Proceedings of the 3rd Workshop on Natural Language
  Processing for Conversational AI}, pages 251--253, Online. Association for
  Computational Linguistics.

\bibitem[{Zhou et~al.(2021{\natexlab{b}})Zhou, Jandaghi, Cho, Lin, Pujara, and
  Ren}]{zhou-etal-2021-probing-commonsense}
Pei Zhou, Pegah Jandaghi, Hyundong Cho, Bill~Yuchen Lin, Jay Pujara, and Xiang
  Ren. 2021{\natexlab{b}}.
\newblock \href {https://doi.org/10.18653/v1/2021.findings-emnlp.349} {Probing
  commonsense explanation in dialogue response generation}.
\newblock In \emph{Findings of the Association for Computational Linguistics:
  EMNLP 2021}, pages 4132--4146, Punta Cana, Dominican Republic. Association
  for Computational Linguistics.

\bibitem[{Zhou et~al.(2020)Zhou, Hu, Zhang, Liang, Sun, Xiong, and
  Tang}]{zhou2020towards}
Wangchunshu Zhou, Jinyi Hu, Hanlin Zhang, Xiaodan Liang, Maosong Sun, Chenyan
  Xiong, and Jian Tang. 2020.
\newblock Towards interpretable natural language understanding with
  explanations as latent variables.
\newblock \emph{Advances in Neural Information Processing Systems},
  33:6803--6814.

\end{thebibliography}
\bibliographystyle{acl_natbib}
\clearpage
\appendix
\section{Additional Experiments}
\label{sec:appendix}
\paragraph{How do models with different architectures and sizes perform at abductive reasoning?}

Table~\ref{tab:arch_size} summarizes the results on the $\alpha$NLI dataset with different model architectures and model sizes, which are obtained from the same grid search described in Sec.~\ref{sec:setup}.
Within the same architecture, models with more parameters are better at abductive reasoning.
When comparing between BART and T5, BART can produce consistent better results at each size.

\paragraph{Does a learnable $p(z|x)$ model lead to better performance?}
Here we test if a learnable $p(z|x)$ model instead of a uniform $p(z|x)$ model leads to better performance.
We should note that a learnable $p(z|x)$ model may result in reasoning shortcuts: because if the signal from $p(z|x)$ is too strong, then this term will dominate Eq.~\ref{bayes}; thus, $p(z|x,y)$ computed in this way is no longer a result of thinking backwards.
We parametrize the learnable $p(z|x)$ model by a BART-large model, which takes $x$ as an input and returns a probability distribution over all sequences.
Table~\ref{tab:analysis} shows the comparison between the two $p(z|x)$ models on the $\alpha$NLI dataset.
Although the uniform $p(z|x)$ model outperforms the learnable $p(z|x)$ model, the difference between them is not significant.

\paragraph{How do methods without plausibility annotations perform in presence of distractors?}
In order to test the robustness of different methods without plausibility annotations, we evaluate them on two types of distractors added to the $\alpha$NLI test set.
The first type of distractor randomly samples a third explanation from another example, and the second type of distractor constructs a third explanation with randomly sampled words from the vocabulary of the $\alpha$NLI dataset with a length that falls in-between the lengths of the two original explanations.
Table~\ref{tab:analysis} compares the results with and without the distractors.
Notice that after adding a third option, the chance of getting the plausible explanation with a random guess is $\frac{1}{3}$.
LiPoR's accuracy drops significantly with the presence of distrators, while the relative decrease for GPT NEO is smaller.
Furthermore, the zero-shot results (i.e., ZS and GPT NEO) suggest that it is more difficult to identify the first type of distractor than the second one.
Our interpretation for a worse performing LiPoR's on distractors is that the distrators break our assumption: $p(z|x)$ is no longer uniform, and the probability of a distracting explanation is independent of the probability of $x$.
Therefore, the original factorization in Eq.~\ref{eq:joint_prob} no longer applies.
To build an unsupervised system that is robust to distractors requires incorporating the new assumptions in the data generating process.

\begin{table}[t]
\centering
\resizebox{0.5\linewidth}{!}{%
\begin{tabular}{lrr}
\toprule
      & \multicolumn{1}{l}{BART} & \multicolumn{1}{l}{T5} \\ \midrule
small & -                        & 54.14                  \\
base  & 60.08                    & 57.31                  \\
large & 71.56                    & 65.48   \\
\bottomrule
\end{tabular}%
}
\caption{Comparison between different model architectures and model sizes on the $\alpha$NLI dataset.}
\label{tab:arch_size}
\end{table}

\begin{table}[t]
\centering
\resizebox{\linewidth}{!}{%
\begin{tabular}{lrrr}
\toprule
                           & \multicolumn{1}{l}{Original} & \multicolumn{1}{l}{+Rand. E's} & \multicolumn{1}{l}{+Rand. W's} \\ \midrule
GPT NEO                    & 57.47                        & 51.12                          & 57.37                          \\
ZS                         & 50.96                        & 34.39                          & 38.22                          \\
\cellcolor[HTML]{FFFFFF}LL & 57.40                         & 53.48                          & 53.52                          \\
LiPoR w/ unif. $p(z|x)$      & \textbf{71.56}                        & 58.58                          & 57.40                           \\
LiPoR w/ learned $p(z|x)$    & 69.92                        & \textbf{59.14}                          & \textbf{59.24}             \\
\bottomrule
\end{tabular}%
}
\caption{Comparison between different unsupervised approaches on the $\alpha$NLI test set. +Rand. E's is adding a random explanation taken from another example. +Rand. W's is adding random words from the vocabulary of $\alpha$NLI whose length is between the lengths of two original explanations. Best results for each setting is in boldface.}
\label{tab:analysis}
\end{table}



\end{document}